\documentclass[conference]{IEEEtran}
\IEEEoverridecommandlockouts
\usepackage{cite}
\usepackage{amsmath,amssymb,amsfonts}
\usepackage{algorithmic}
\usepackage{graphicx}
\usepackage{textcomp}
\usepackage{xcolor}
\usepackage{multirow}
\usepackage{subfigure} 
\def\BibTeX{{\rm B\kern-.05em{\sc i\kern-.025em b}\kern-.08em
    T\kern-.1667em\lower.7ex\hbox{E}\kern-.125emX}}
\begin{document}

\title{A Light-weight CNN Model for Efficient Parkinson's Disease Diagnostics\\
}

\author{\IEEEauthorblockN{Xuechao Wang\IEEEauthorrefmark{1},
Junqing Huang\IEEEauthorrefmark{2}, 
Marianna Chatzakou\IEEEauthorrefmark{3},\\
Kadri Medijainen\IEEEauthorrefmark{4},
Pille Taba\IEEEauthorrefmark{5},
Aaro Toomela\IEEEauthorrefmark{6},\\
Sven N{\~omm}\IEEEauthorrefmark{7},
Michael Ruzhansky \IEEEauthorrefmark{8}}

\IEEEauthorblockA{\IEEEauthorrefmark{1} \IEEEauthorrefmark{2} \IEEEauthorrefmark{3} \IEEEauthorrefmark{8} Department of Mathematics: Analysis, Logic and Discrete Mathematics, Ghent University, Ghent, Belgium\\
Email: \IEEEauthorrefmark{1}xuechao.wang@ugent.be,
\IEEEauthorrefmark{2}junqing.huang@ugent.be,
\IEEEauthorrefmark{3}marianna.chatzakou@ugent.be,
\IEEEauthorrefmark{8}michael.ruzhansky@ugent.be,
}

\IEEEauthorblockA{\IEEEauthorrefmark{8} School of Mathematical Sciences, Queen Mary University of London, Mile End Road, London E1 4NS, United Kingdom}

\IEEEauthorblockA{\IEEEauthorrefmark{7} Department of Software Science, Faculty of Information Technology, Tallinn University of Technology, \\Akadeemia tee 15 a, 12618, Tallinn, Estonia \\
Email: \IEEEauthorrefmark{7}sven.nomm@ttu.ee}

\IEEEauthorblockA{\IEEEauthorrefmark{4} Institute of Sport Sciences and Physiotherapy, University of Tartu, Puusepa 8, Tartu 51014, Estonia \\
E-mail: \IEEEauthorrefmark{4} kadri.medijainen@ut.ee}

\IEEEauthorblockA{\IEEEauthorrefmark{5} Department of Neurology and Neurosurgery, University of Tartu, Puusepa 8, Tartu 51014, Estonia 
}
\IEEEauthorblockA{\IEEEauthorrefmark{5} Neurology Clinic, Tartu University Hospital, Puusepa 8, Tartu 51014, Estonia \\
E-mail: \IEEEauthorrefmark{5} pille.taba@kliinikum.ee}

\IEEEauthorblockA{\IEEEauthorrefmark{6} School of Natural Sciences and Health, Tallinn University,
Narva mnt. 25, 10120, Tallinn, Estonia\\
E-mail: \IEEEauthorrefmark{6} aaro.toomela@tlu.ee}
}

\maketitle

\begin{abstract}
In recent years, deep learning methods have achieved great success in various fields due to their strong performance in practical applications. In this paper, we present a light-weight neural network for Parkinson's disease diagnostics, in which a series of hand-drawn data are collected to distinguish Parkinson's disease patients from healthy control subjects. The proposed model consists of a convolution neural network (CNN) cascading to long-short-term memory (LSTM) to adapt the characteristics of collected time-series signals. To make full use of their advantages, a multilayered LSTM model is firstly used to enrich features which are then concatenated with raw data and fed into a shallow one-dimensional (1D) CNN model for efficient classification. Experimental results show that the proposed model achieves a high-quality diagnostic result over multiple evaluation metrics with much fewer parameters and operations, outperforming conventional methods such as support vector machine (SVM), random forest (RF), lightgbm (LGB) and CNN-based methods.
\end{abstract}


\section{Introduction}
Parkinson's disease (PD) is a neurodegenerative disease that is the second most common neurological disorder after Alzheimer's disease \cite{calne1993treatment}. From $1990$ to $2016$, the estimated global population affected by PD increased more than double (from $2.5$ million to $6.1$ million), affecting $1$-$2$ in every $1,000$ people over $60$ years old \cite{dorsey2018global}. PD symptoms such as bradykinesia, tremors, and rigidity can severely affect the patient's quality of life \cite{louis2015tremor}, family relationships, and social functioning, placing a heavy financial burden on individuals and society. Although PD is currently incurable, early diagnosis and proper treatment can allow reliving most of the symptoms. Although some current clinical methods (e.g., optical coherence tomography, magnetic resonance imaging) can help diagnose early PD, they are too harsh and expensive to use, which can lead to patients missing the optimal treatment time. Therefore, researchers from different fields aim to combine knowledge to help people with limited access to medical care have an efficient diagnosis of PD. 

\begin{figure}[t]
	\centering  
	\includegraphics[width=0.5\textwidth ]{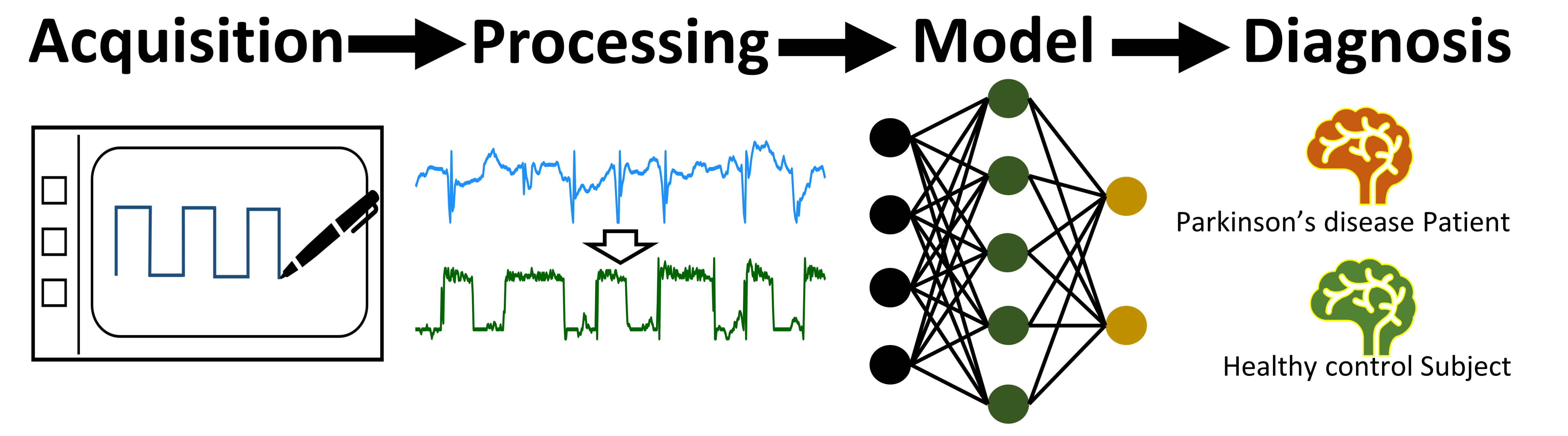}
	\caption{The proposed workflow for Parkinson's disease diagnostics.}
	\label{fig:workflow}
\end{figure}

Hand drawing tests have been used for almost a century as a convenient way to diagnose PD \cite{goodenough1926measurement, saunders2008validity, broderick2009hypometria}. In fact, hand drawing is a complex activity that involves sensorimotor and cognitive components, and changes in it can be considered a promising biomarker for the evaluation of early PD \cite{drotar2013new, rosenblum2013handwriting}.  Originally, hand-drawing tests were performed on paper and pen and analysed by the practitioner. This classical setting has two drawbacks, the first is the subjective component introduced by the human practitioner, and the second is that the naked eye could not capture all the parameters of the movements. The work of \cite{Marquardt199439} that proposed the use of a digital table opened a new research direction with the aim of addressing both drawbacks. With the proliferation of digital devices, it is now possible to record measurement sequences for hand-drawing tests, provided by tablet and pen devices. The machine can capture the parameters invisible to the human eye, and the digital description of the drawing tests is free of subjective components. Initially, the number of parameters analysed was five, while in \cite{DROTAR2016} and \cite{VALLA2022103551} the authors increased this number to hundreds. Furthermore, standard procedures were also described in \cite{Smits2014}. Evidently, the development of hand-drawing-based decision is a noninvasive, real-time, and low-cost solution to support standard clinical assessments by human experts.

\begin{figure}[!t]
	\centering  
	\includegraphics[width=0.5\textwidth]{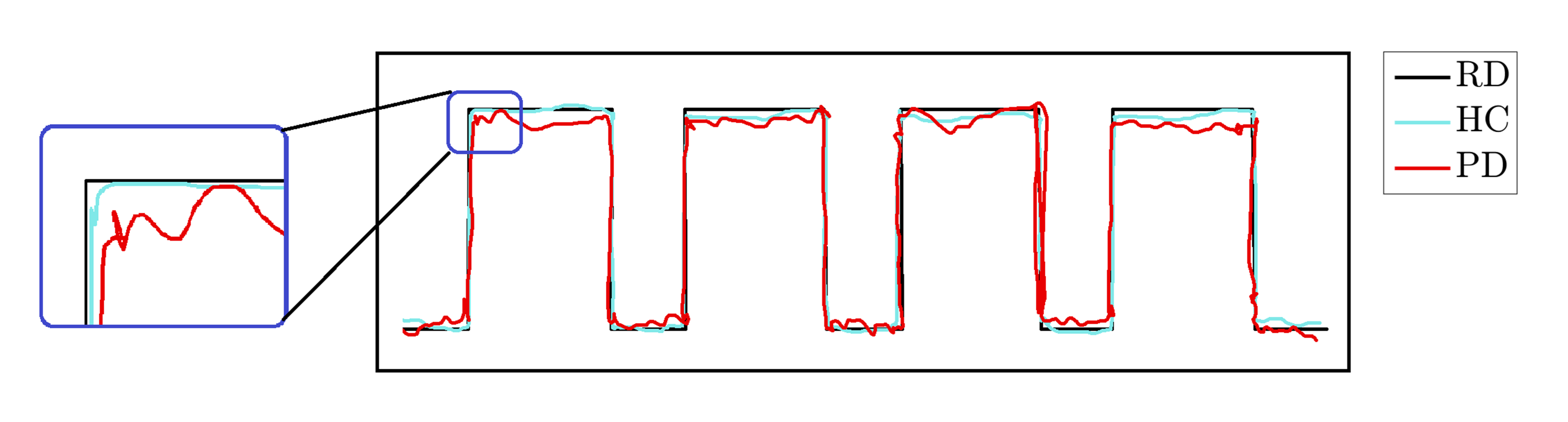}
	\caption{Comparison of the hand drawings of Parkinson's disease (PD) patient and healthy control (HC) subject under a reference drawing (RD).}
	\label{fig:drawing curves}
\end{figure}

Deep learning (DL) based approaches have been gaining great attention and popularity in the analysis of drawing and writing tests. The research reported in \cite{pereira2016deep}, introduced NewHandPD, a data set of signals extracted from an electronic smart pen, and proposed to pose the problem of distinguishing PD from healthy control (HC) as an image recognition task through CNN. The potential of handwritten visual properties to predict PD was investigated by \cite{moetesum2019assessing}. In \cite{nomm2020deep}, the drawn curves were enhanced with kinematic and pressure characteristics to classify the generated images using CNN. \cite{che2017rnn} developed a deep RNN model to learn patient similarity in PD that dynamically matched the temporal patterns of patient sequences to learn patient similarity directly from longitudinal and multimodal patient records. RNN with hyperbolic secant in the gates was investigated in \cite{fujita2021performance} to address the same problem. Although the above models could generally diagnose PD, and the classification performance of diagnostic models has been shown to be comparable to that of human practitioners \cite{zarembo2021cnn}, the key step is to select appropriate characteristics through complex processing, and the overall structure of the network is relatively complex.

To address the shortcomings of the above methods, a simple and efficient deep learning method was proposed for the diagnosis of PD. As shown in Fig.\ref{fig:workflow}, subtle tremor signals were captured from digital Luria alternating series tests (dLAST) collected by the devices to directly reflect the degree of tremor at adjacent moments when performing fine movements. Meanwhile, a lightweight neural network combining LSTM and CNN, called the LSTM-CNN model, was proposed to learn the temporal and spatial characteristics of each tremor signal and thus effectively diagnose early PD. Furthermore, dLAST could not be obtained in the quantity necessary for deep neural network training, and the application of data segmentation technology alleviated the situation of small data sets to some extent. Our extensive experimental results demonstrate that even with much fewer parameters and operations, our method achieves more than $90\%$ on multiple evaluation metrics, delivering strong performance on par with or even better than the state-of-the-art.

The paper is organised as follows. Section \ref{sec:dataset} provides the reader with the necessary information about the data. Section \ref{sec:methodology} presents the research methods and the experimental setting. Section \ref{sec:results} presents the main results of the current studies. Discussion of the results achieved, the limitations of the proposed methods, and possible future directions constitute Section \ref{sec:discussion}.

\section{Material} \label{sec:dataset}

In this research, a data set is considered. The data set, here and later referred to as DraWritePD, was acquired from $49$ participants, with a mean age of $74.1$ years and a similar gender distribution. Within the group of patients with PD, the age deviation was approximately $3.35$ years, while within the group of subjects with HC, the age deviation was $4.55$ years, making both groups very similar. 

Data acquisition was performed with an iPad Pro $9.7$ inch ($2016$) equipped with an Apple Pencil. As shown in Fig.\ref{fig:drawing curves}, participants were asked to mimic the reference pattern to draw. During this process, the iPad Pro scanned the Apple Pencil signal at $240$ points per second. As shown in Table \ref{tab:dataset description}, for each scan, the device captures six time sequence parameters: azimuth ($a$); altitude ($l$); pressure ($p$); timestamp ($t$); x-Axis ($x$); y-Axis ($y$).

\renewcommand{\arraystretch}{1.5}
\begin{table}[ht] 
	\centering
	\scriptsize 
	\caption{Description of the DraWritePD dataset}
	\setlength{\tabcolsep}{1.5mm}{
	\begin{tabular}{| c || c| c | c| c | c | c|}
		\hline 
		Parameter & azimuth &	altitude & pressure & timestamp  & x-Axis & y-Axis\\
		\hline
		Notation & $a$ & $l$ & $p$ & $t$ & $x$ & $y$\\
		\hline
		Quantitation & rad & rad & psi & sec & mm & mm\\
		\hline
	\end{tabular}}
	\label{tab:dataset description}
\end{table}

Originally, the dLAST consisted of $3$ tests: \textit{$\Pi$}\textit{$\Lambda$}, \textit{$\Pi$} and \textit{sin wave}, each with $3$ exercises: \textit{continue}, \textit{copy} and \textit{trace} applied to each test. The general task was to complete a set of tests consisting of $9$ different drawing exercises. To optimise the testing procedure within the frameworks of the present contribution, only the \textit{$\Pi$} test was considered. Specifically, the graphical plots of \textit{$\Pi$} test are demonstrated in Fig. \ref{fig:drawing curves}.

The data acquisition process was carried out under strict privacy law guidance. The Research Ethics Committee approved the study of the University of Tartu (No.$1275T-9$).

\section{Methodology} \label{sec:methodology}

In this section, we illustrate the proposed LSTM-CNN model for the diagnosis of PD. The framework starts with preprocessing methods for standardising the data.  Regarding the model architecture, we propose a lightweight hybrid model that is composed of an LSTM block cascaded with a CNN-based classifier. We will demonstrate in detail the performance and efficiency of the proposed LSTM-CNN model.

\begin{figure}[!t]
	\centering  
	\includegraphics[width=0.5\textwidth ]{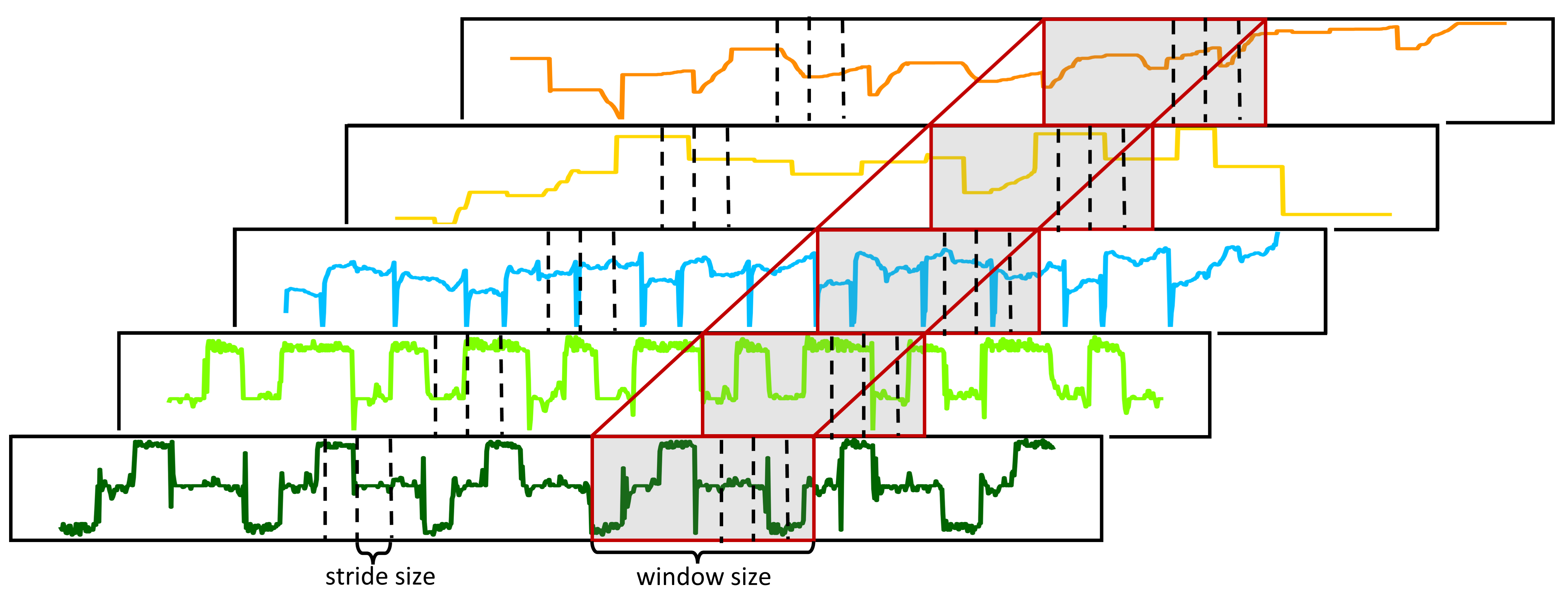}
	\caption{The scheme of data segmentation with temporal overlap region used in the framework. The \emph{window size} and \emph{stride size} control the length and overlap of the patch.}  
	\label{fig:segmentation}
\end{figure}

\subsection{Data pre-processing}

In many practical applications, the raw data may not be perfectly collected --- for example, some data may be missing, or inevitably contain some abnormal features that are not appropriate for direct use in training a regression or a classification model. In this situation, it is always necessary to use some pre-processing methods to standardise the data or improve their quality. Considering the characteristics of the DraWritePD data set, we employ the following preprocessing methods to ensure that our data are suitable for training a diagnostic model.

\textbf{Normalisation} 
 As explained in Section \ref{sec:dataset}, the DraWritePD data set includes parameters with different ranges. It is necessary to rescale or normalise the data. The Min-Max normalisation technique is used to linearly convert each individual parameter signal in the range from $0$ to $1$. 
 
\begin{figure}[t]
    \centering
    \subfigure[]{
        \includegraphics[width=0.2\textwidth]{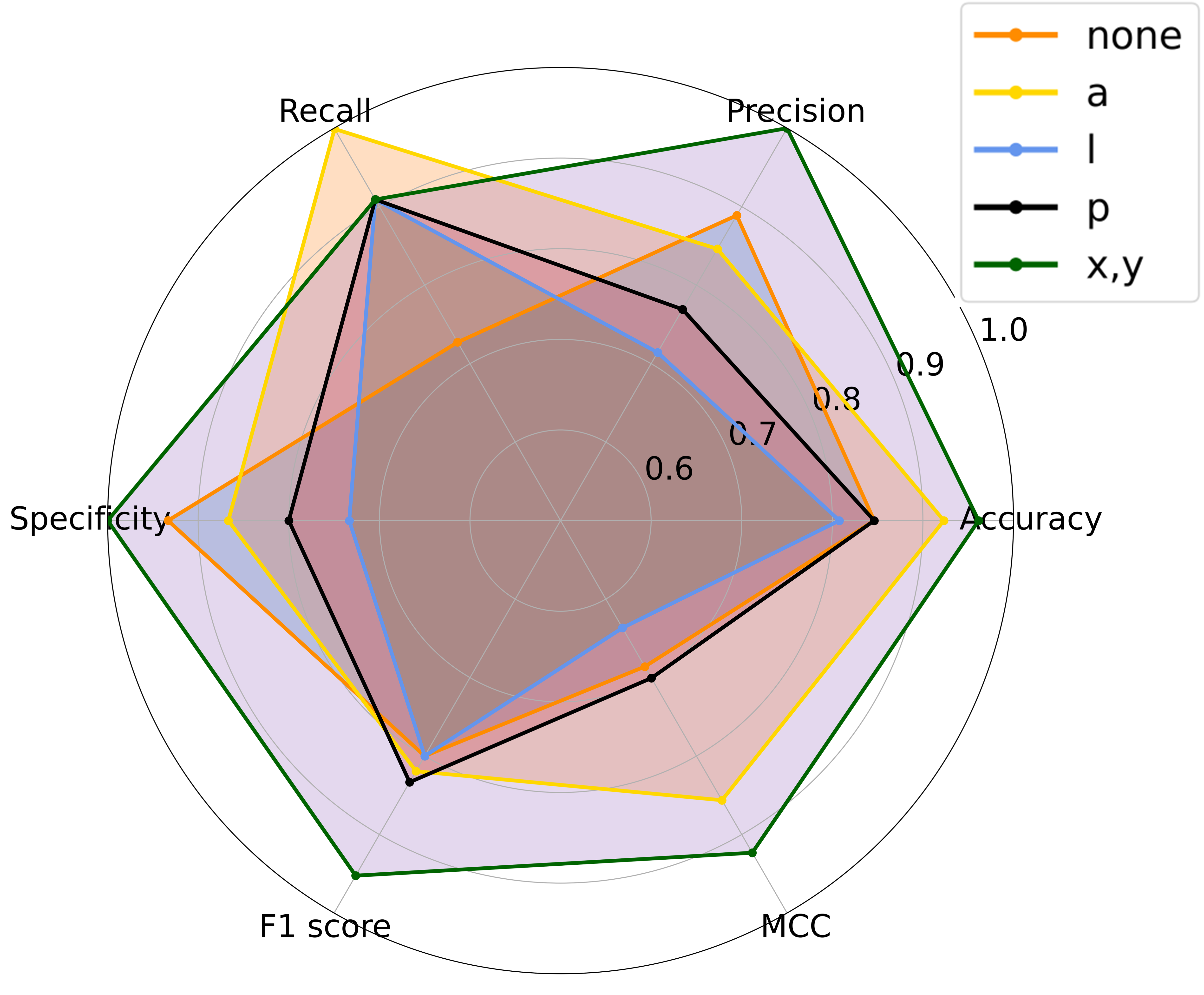}
    }
    \subfigure[]{
        \includegraphics[width=0.2\textwidth]{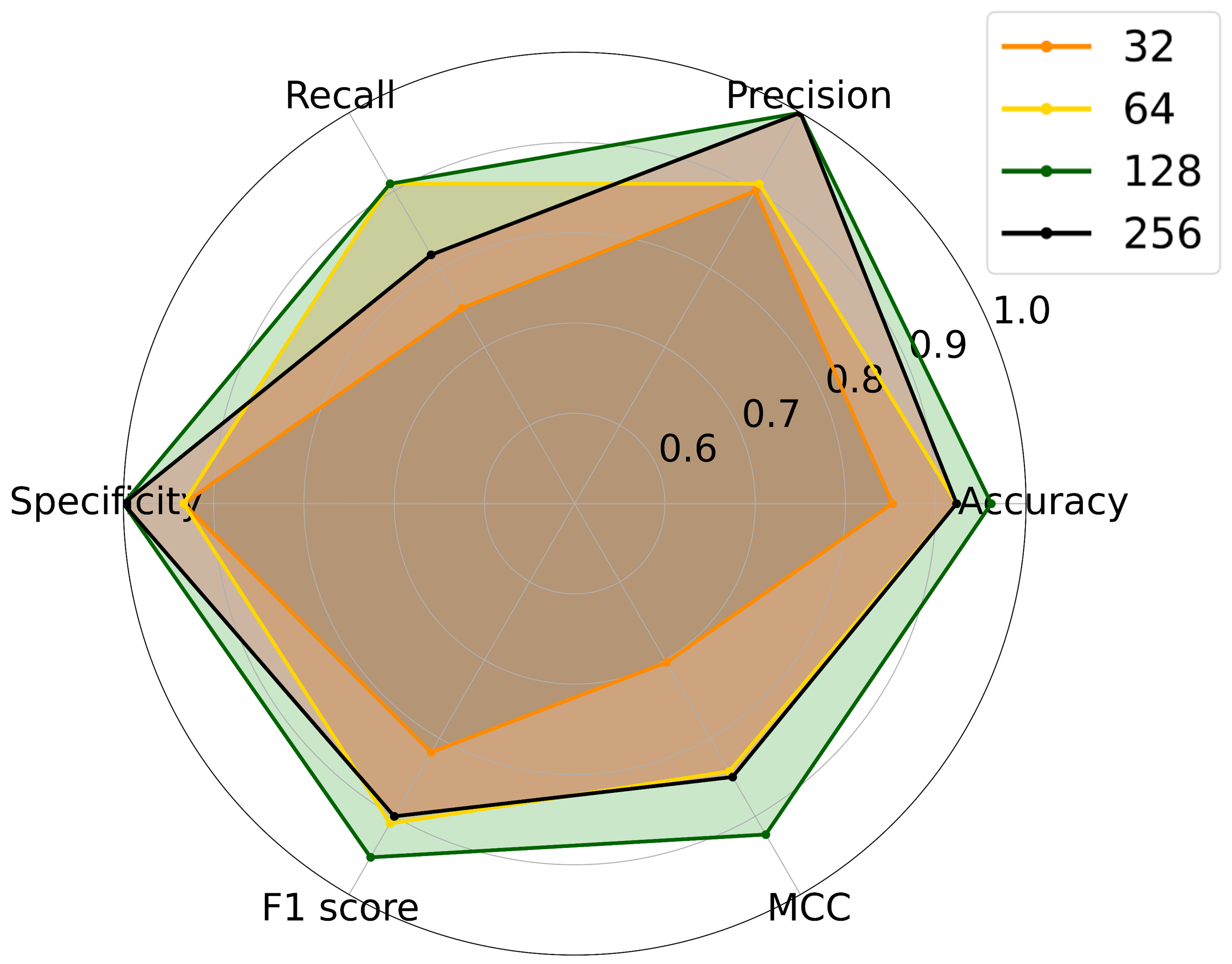}
    }
    \caption{The effect of hyper-parameters in data pre-processing on the model, where (a) is the effect of transforming different raw parameter signals into velocity features in Feature engineering and selection on the model, and (b) is the effect of \textit{window size} in Segmentation on the model.} 
	\label{fig:parameter}
\end{figure}

\textbf{Feature engineering and selection}
After the normalising procedure, the feature engineering process proposed in \cite{DROTAR2016} suggests considering the kinematic parameters of the movements of the tip of the pen. Kinematic parameters of fine motor movements (observed during writing and drawing activities) would reflect tremor, freezing, and other symptoms caused by progressing PD \cite{rosenblum2013handwriting}, and we have demonstrated this conclusion through experiments (see Fig. \ref{fig:parameter} (a)). Within the framework of the present studies, main attention is paid to the projections of the velocity of x- and y-coordinate parameter signals on the coordinate axis given by Eq.(1), while keeping the other parameters constant.
\begin{equation}
v_x^t= (x_t-x_{t-1})/\delta_t, \quad  v_y^t= (y_t-y_{t-1})/\delta_t,
\end{equation}
where ($x_{t-1}$,$y_{t-1}$) and ($x_t$,$y_t$) are the x- and y-coordinate position information of two adjacent time points, and $\delta_t$ is the time interval between two contiguous sampling data points. The use of velocity also has the advantage of converting the non-stationary $x,y$ coordinate-based features into stationary ones, leading to a more tractable classification under the proposed model. 

\textbf{Segmentation} 
Afterwards, data segmentation is adopted in the proposed framework to generate suitable data samples for model training. The approach of using a sliding window to randomly select a local patch is frequently used in time-series classification and has been demonstrated to be useful in improving model performance. Fig. \ref{fig:segmentation} shows an example based on the DraWritePD data set. The multichannel time-series data are cropped into small patches with a slight overlap to preserve the temporal information. The parameters \textit{window size} ($w$) and \textit{stride size} ($s$) control the length and overlap of the resampling patches. The choice of \textit{window size} and \textit{stride size} depends on the concrete applications. Experiments illustrated by Fig. \ref{fig:parameter} (b) demonstrate that the \textit{$w$=128} provides balance between the performance and the efficiency under the proposed LSTM-CNN architecture. It is also worth noting that such segmentation generates more data samples for training, which is crucial for limited or small data sampling cases, including the proposed method based on the LSTM-CNN deep learning architecture.

\subsection{Model Structure}

\begin{figure*}[!t]
	\centering  
	\includegraphics[width=\textwidth ]{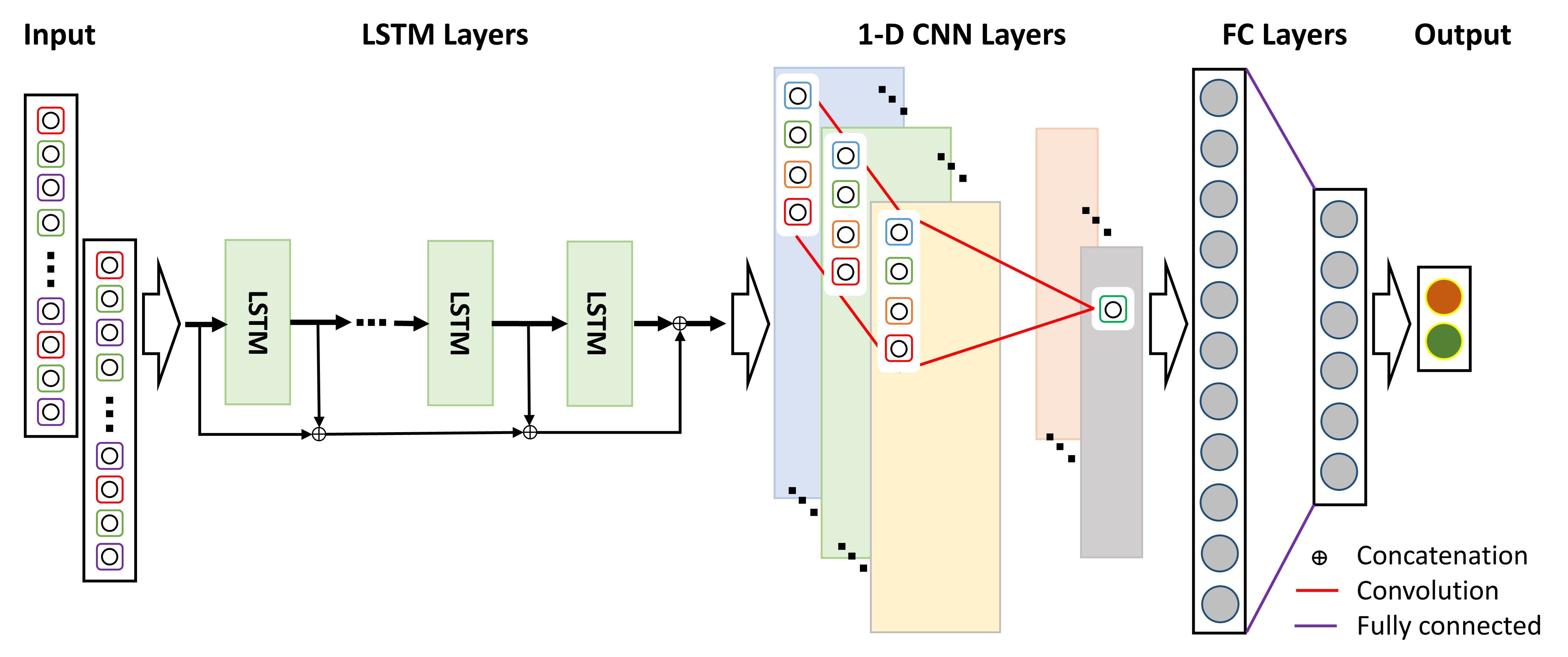}
	\caption{The network structure of the proposed LSTM-CNN model.} 
	\label{fig:lstmcnn}
\end{figure*}

The proposed model structure is a hybrid of the LSTM and CNN models. Such a combination helps to explore the advantages of both the LSTM and CNN models. In this LSTM-CNN model, convolution operators are also reduced to $1$D instances to reduce the computational cost. For clarity, we introduce the model in Fig. \ref{fig:lstmcnn} according to the characteristics of the LSTM and CNN blocks.

\textbf{LSTM Block} 
The LSTM unit is a subtype of the recurrent neural network (RNN). It was originally proposed as an efficient and scalable building block for analysing complex sequential data or time-series data. The LSTM unit contains a group of special memory cells and is capable of extracting temporal features of data based on the memory of historical information, giving a great advantage over CNN in the extraction of sequence data. For each memory cell, the input data is first sent to different gates, including the input gate, the forget gate, and the output gate, to control the behaviour of each memory cell, and then the output is sent as input at the corresponding moment of the next memory cell. For more details on the structure and properties of LSTM, we refer to \cite{hochreiter1997long}. In our LSTM-CNN model, the LSTM block consists of a series of cascaded LSTM units. Experiments show that, in order to improve efficiency while maintaining model performance, we can choose only one LSTM unit to construct the LSTM block. Furthermore, the output of each LSTM unit is concatenated with the original input data and sent to the subsequent CNN block for robust classification, in which the LSTM output is dimension-expanded (represented as batch size, $1$, window size, and feature dimension) to accommodate the shape of the input convolutional layer. 

\textbf{CNN Block} 
The CNN unit is suitable for classification or recognition tasks due to its ability to learn discriminate representations. We explore the performance of convolutional networks and follow the architecture and suggestions in \cite{krizhevsky2012imagenet}. As shown in Fig. \ref{fig:lstmcnn}, the proposed model contains two CNN units, where each consists of a $1$D temporal convolutional layer, a rectified linear unit (ReLU) layer and a 1D max pool layer. Specifically, the $1$D convolutional layer is the most important component due to its unique feature extraction ability, where in the first convolution unit, the $16$ convolution kernels are used for feature extraction, and in the second unit, the $32$ convolution kernels are used for deeper feature extraction operations on the output of the feature by the upper layer. The size of each convolution kernel is $3\times1$, and the sliding step size of the convolution window is fixed to $2$ throughout all experiments. Next, a ReLU layer is used to activate its output. A maximum pool layer is used after the ReLU layer to perform the downsampling operation and reduce parameters while maintaining dominant features, with a sampling kernel size of $2\times1$ and a stride size of $2$. Notice that all operators are reduced to cases of $1$ -D for efficiency.

\textbf{FC Block}
The fully connected (FC) block --- consisting of a fully connected layer, a ReLU activation layer and a dropout layer --- is used in the proposed model. During the training phase, the dropout layer temporarily removes nodes from the network with a probability of $0.5$. For stochastic gradient descent, since it is randomly dropped, each mini-batch is training a different network to prevent model overfitting and improve model performance. Furthermore, a fully connected layer is deployed after the dropout layer to convert the previous output, the value of which represents the probabilities belonging to each class.

\section{Experimental Results}\label{sec:results}

In this section, we evaluate and analyse the performance of the proposed LSTM-CNN model using the DraWritePD data set.  The model runs on the desktop PC with an Intel(R) Core(TM) $3.60$ GHz($8$ CPU), $32$GB RAM, and an NVIDIA RTX$3070$Ti GPU with $8$ GB memory. 

\subsection{Dataset}

The DraWritePD set contains $157$ pieces of sequence data from $29$ subjects with HC and $20$ patients with PD. The raw sequence data needs to be cropped into patches before being fed to the LSTM-CNN model. The class imbalance problem may occur during the segmentation due to the different lengths of the sequence data. As shown in Fig. \ref{fig:segmentation}, a nonuniform sampling strategy with varying stride size is adopted to impose the number of generated patch data in each class to be the same. The statistics of the training and testing dataset is listed in Table \ref{tab:dataste}.

\renewcommand{\arraystretch}{1.5}
\begin{table}[ht]
	\centering
	\scriptsize 
	\caption{The information of training set and testing set.}
    \setlength{\tabcolsep}{4mm} {
	\begin{tabular}{|c||c|c|c|c|}
		\hline 
	    \multirow{2}*{}  &  \multicolumn{2}{c|}{Training set} & \multicolumn{2}{c|}{Testing set} \\
	    \cline{2-5}
		                 &       HC       &         PD        &         HC       &       PD      \\
		\hline
		\hline
		Participant     &       25       &         16        &         4        &        4      \\
		\hline
		Sequence set (S)      &       80       &         51        &         15       &       11      \\
		\hline
		Patch set (P)      &     16166      &       16836       &       3670       &     3319      \\
		\hline
	\end{tabular}}
	\label{tab:dataste}
\end{table}

During the training phase, the patch data set was randomly divided in the $8:2$ ratio into a training patch data set and a validation patch data set. During the testing phase, the proposed model was first evaluated based on the testing patch data set and then applied to the testing sequence data set, where the predicted result of each raw sequence data set was determined by the majority vote of the prediction result of the patch data. To clarify, we denote the patch data set testing as $P$ and the raw sequence data set as $S$, and independently evaluate the performance of the proposed model on the two cases.

\subsection{Experimental Setup}
In order to fully exploit the performance of the proposed LSTM-CNN model, we use a cross-validation strategy to optimally choose the parameters. Adam \cite{kingma2014adam} optimiser is used to train the model, and the initial learning rate is set to $0.001$. Furthermore, the cross-entropy loss function is used for model fitting and the batch size is set to $64$. The proposed model is completed in $200$ epochs with the loss curve shown in Fig. \ref{fig:train_val_curve}. We use the metrics: accuracy, precision, recall, specificity, $F_1$ score, and Matthews correlation coefficient (MCC) for evaluation, where the latter has been adopted by many existing methods to describe the different aspects of the performance of a classifier \cite{baldi2000assessing}. Once the training phase is completed, the one with the best fitness value is chosen for testing. Moreover, the length of segmented patches and the choice of feature selection are also discussed to interpret their roles in determining the model performance. As shown in Fig.~\ref{fig:parameter}, the model achieves the best classification result when the window size is $128$ by using the ($v_x,v_y$) velocity characteristics. The model performance is eventually tested on both the original sequences dataset ($S$) and the segmented patches dataset ($P$).

\subsection{Quantitative Evaluation and Comparison}

\renewcommand{\arraystretch}{1.5}
\begin{table*}[ht]
	\centering
	\scriptsize 
	\caption{Quantitative comparison of different classification methods.}
 \setlength{\tabcolsep}{3mm}{
	\begin{tabular}{ | c || c || c | c | c | c | c | c |}
  \hline
  \multirow{2}*{Model} & \multirow{2}*{Inference time (s)} &  \multicolumn{6}{c|}{Metric} \\
	\cline{3-8}
		~ & ~ &  Accuracy (P/S) & Precision (P/S) & Recall (P/S) & Specificity (P/S) & $F_1$ score (P/S) & MCC (P/S) \\
		\hline
        \hline
        LR & 0.034 & 0.8061 / 0.9231  & 0.8559 / 0.8462 & 0.8565 / \textbf{1.00} & 0.7018 / 0.8667 & 0.8562 / 0.9167 & 0.5585 / 0.8563  \\
        \hline
        SVM & 6.060 &  0.8371 / 0.8846  & 0.8657 / 0.7857 & 0.8977 / \textbf{1.00} & 0.7119 / 0.8000 & 0.8814 / 0.8800 & 0.6229 / 0.7928  \\
        \hline
        RF & 9.526 &  0.8339 / 0.8462  & 0.9015 / 0.8889 & 0.8412 / 0.7273 & 0.8088 / 0.9333 & 0.8729 / 0.8000 & 0.6368 / 0.6860  \\
        \hline
        LGB & 0.161 & 0.7889 / 0.8077   & 0.9183 / 0.8750 & 0.7538 / 0.6364 & 0.8613 / 0.9333 & 0.8280 / 0.7368 & 0.5800 / 0.6098  \\
        \hline
        
        MLP & 4.095 & 0.8274 / 0.8846  & 0.8598 / 0.8333 & 0.8921 / 0.9091 & 0.6891 / 0.8667 & 0.8756 / 0.8696 & 0.5950 / 0.7688  \\
        \hline
        AlexNet & 4.143 &  0.7872 / 0.8846  & 0.9093 / \textbf{1.00} & 0.7606 / 0.7273 & 0.8425 / \textbf{1.00} & 0.8284 / 0.8421 & 0.5662 / 0.7785  \\
        \hline
        \hline
        LSTM-CNN (Ours) & 4.212 &  0.7994 / \textbf{0.9615}  & 0.8883 / \textbf{1.00} & 0.8039 / 0.9091 & 0.7900 / \textbf{1.00} & 0.8440 / \textbf{0.9524} & 0.5720 / \textbf{0.9232}  \\
        \hline     
	\end{tabular}}
	\label{tab:model structure}
\end{table*}

We provide a quantitative comparison to demonstrate the effectiveness and advantages of the proposed LSTM-CNN model. First, we compare it with some traditional machine learning (ML)-based classifiers, which include the Logistic Regression (LR), Support Vector Machine (SVM), Random Forest (RF), and LightGBM (LGB) \cite{ke2017lightgbm}. For each classifier, we take $10$-fold cross-validations and the grid search algorithm to optimise the parameters and to ensure the robustness of the results. The training and validation of all these ML classifiers are run under Python's scikit-learn library \cite{pedregosa2011scikit}. As shown in Table \ref{tab:model structure}, our model has obvious advantages in most classification metrics,  in terms of accuracy increased by $3.8$\%, $F_1$ score increased by $3.5$\%, and MCC increased by $6.6$\%. Regarding the efficiency of the model, our method outperforms $1.8$ seconds and $5.3$ seconds compared to the SVM model and the RF model, although it is slower than the optimised LR model and the LGB model.

\begin{figure}[t]
    \centering
    \subfigure[]{
        \includegraphics[width=0.22\textwidth]{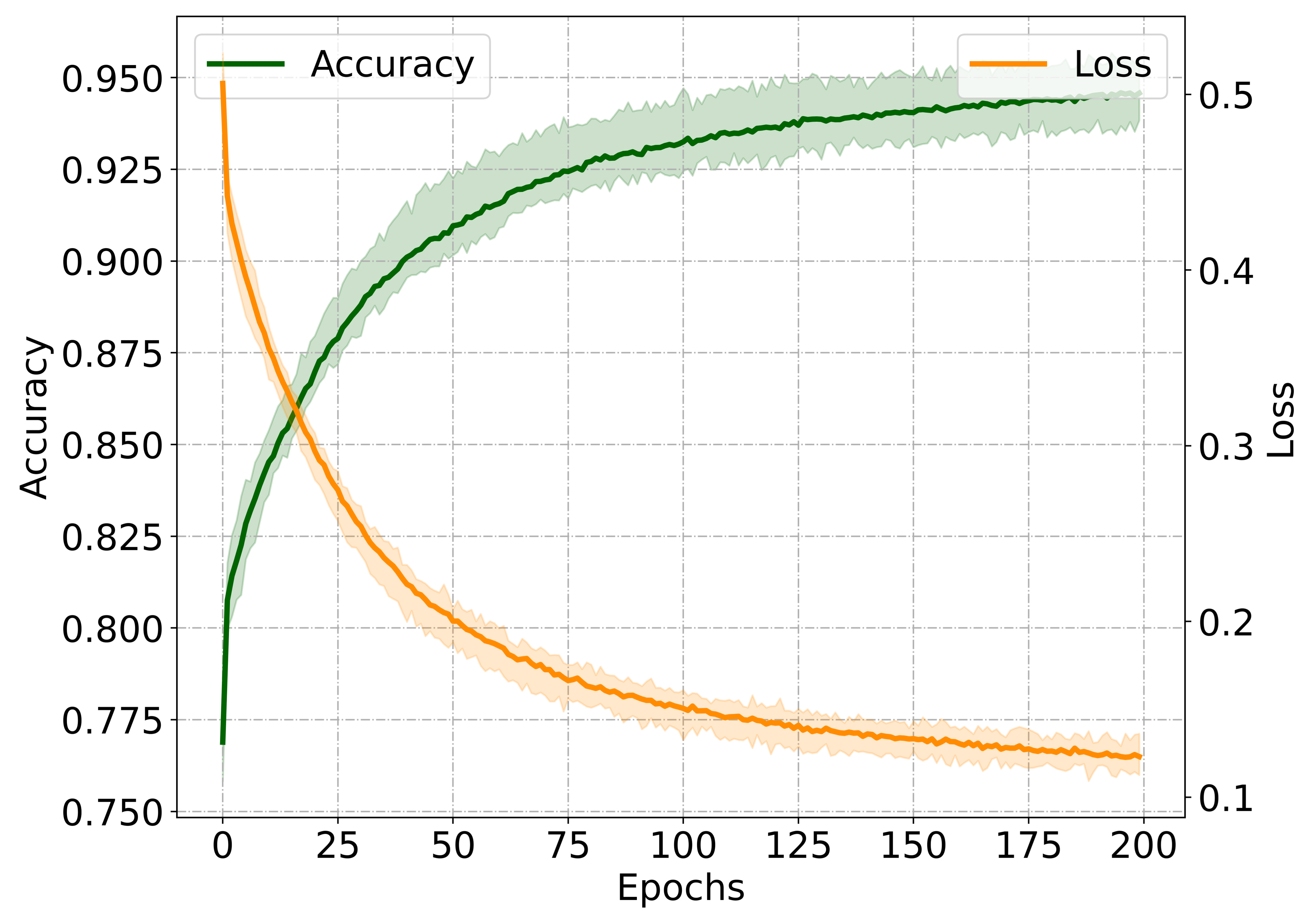}
        }
    \subfigure[]{
        \includegraphics[width=0.22\textwidth]{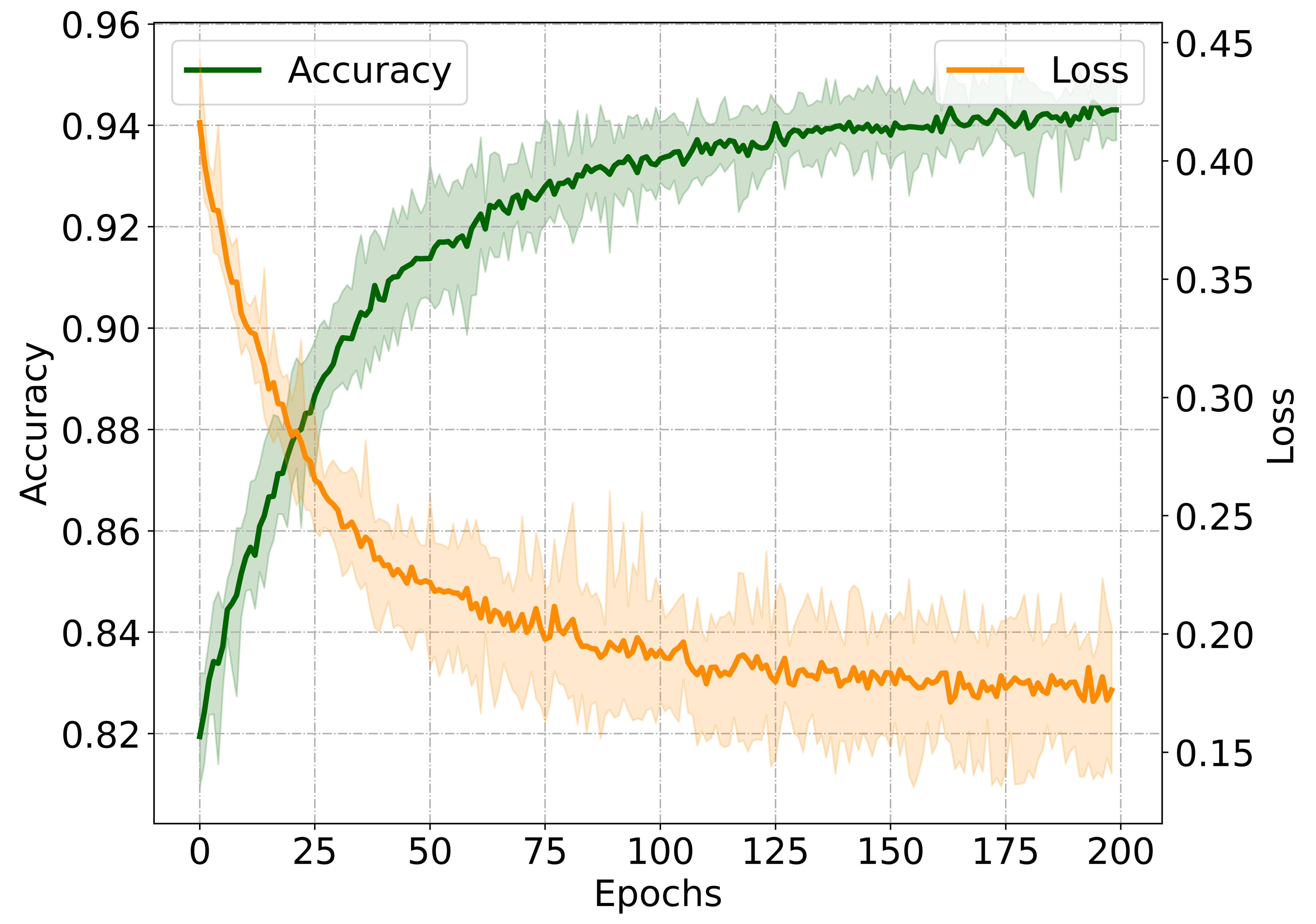}
    }
    \caption{The accuracy and loss curves of the LSTM-CNN model on the training set(a) and the validation set(b), respectively, where the solid curve represents the average of multiple experiments, and the shaded part represents the range of the results of multiple experiments (N=$10$).} 
	\label{fig:train_val_curve}
\end{figure}

Additionally, we also compare the performance of different neural network models on this task. The multilayer perceptron (MLP) is the basic model and its structure consists of two fully connected layers. AlexNet adopts a convolutional neural network structure similar to AlexNet\cite{krizhevsky2012imagenet}, but to adapt to the size of the input data, the internal parameters are modified. In our work, to improve the efficiency of the LSTM-CNN, in addition to the $1$D convolution operation in the CNN block, an LSTM block is added, which contains a concatenation operation. Additionally, as shown in Fig.\ref{fig:train_val_curve}, the average of multiple experimental results (N=$10$) is used as the model result. Finally, let us point out that our methods achieve optimal results in all metrics with the accuracy rate being $96.15\%$, the $F_1$ score being $95.24\%$, and the MCC being $92.32\%$. Specifically, in the $S$ testing set, only one sequence data from the PD category is misclassified, and the remaining 25 sequence data are correctly classified.

In summary, the model proposed in this article could not only achieve high recognition accuracy, but also significantly simplify the structure of the model and improve the efficiency of deep learning models in the diagnosis of PD.

\section{Conclusions}\label{sec:discussion}
This article proposes a lightweight deep neural network that combines convolutional layers $1$D with LSTM for the diagnosis of PD based on hand-drawn tests. In particular, the concatenation operation is added after the LSTM layer, which significantly improves the goodness of the model. Not only can it avoid complex feature extraction, but also it has a high recognition accuracy under the premise of a few model parameters.

\section*{Acknowledgements}
This study was supported by the PRG Grant$957$ of the Estonian Research Council. This work in the project ``ICT programme'' was supported by the European Union through the European Social Fund. It was also partially supported by the FWO Odysseus 1 grant G.0H94.18N: Analysis and Partial Differential Equations, and the Methusalem programme of the Ghent University Special Research Fund (BOF) (Grant number 01M01021). Michael Ruzhansky is also supported by EPSRC grant EP/R003025/2. Marianna Chatzakou is a postdoctoral fellow of the Research Foundation – Flanders (FWO)  under the postdoctoral grant No 12B1223N.
\bibliography{ref}

\end{document}